\documentclass[lettersize,journal]{IEEEtran}
\usepackage{amsmath,amsfonts}
\usepackage{algorithmic}
\usepackage{algorithm}
\usepackage{array}
\usepackage[caption=false,font=normalsize,labelfont=sf,textfont=sf]{subfig}
\usepackage{textcomp}
\usepackage{stfloats}
\usepackage{url}
\usepackage{verbatim}
\usepackage{graphicx}
\usepackage{cite}

\usepackage{balance} 
\usepackage{booktabs}
\usepackage{multirow}
\usepackage{adjustbox}
\usepackage{enumitem}
\usepackage{balance}

\usepackage{hyperref}

\begin{document}

\title{Self-Awareness Safety of Deep Reinforcement Learning in Road Traffic Junction Driving}

\author{Zehong~Cao,~\IEEEmembership{}
        Jie Yun~\IEEEmembership{}


\thanks{Z. Cao is with STEM-AI, University of South Australia, Adelaide, Australia. He was with the School of ICT, University of Tasmania, Hobart, and Australian AI Institute, School of Computer Science, University of Technology Sydney, Sydney, Australia. (E-mail: Jimmy.Cao@unisa.edu.au)}%
\thanks{J. Yun is with School of ICT, University of Tasmania, Hobart, Australia.}%

}

\markboth{  }%
{Cao \MakeLowercase{\textit{et al.}}}


\maketitle

\begin{abstract}

Autonomous driving has been at the forefront of public interest, and a pivotal debate to widespread concerns is safety in the transportation system. Deep reinforcement learning (DRL) has been applied to autonomous driving to provide solutions for obstacle avoidance. However, in a road traffic junction scenario, the vehicle typically receives partial observations from the transportation environment, while DRL needs to rely on long-term rewards to train a reliable model by maximising the cumulative rewards, which may take the risk when exploring new actions and returning either a positive reward or a penalty in the case of collisions. Although safety concerns are usually considered in the design of a reward function, they are not fully considered as the critical metric to directly evaluate the effectiveness of DRL algorithms in autonomous driving. In this study, we evaluated the safety performance of three baseline DRL models (DQN, A2C, and PPO) and proposed a self-awareness module from an attention mechanism for DRL to improve the safety evaluation for an anomalous vehicle in a complex road traffic junction environment, such as intersection and roundabout scenarios, based on four metrics: collision rate, success rate, freezing rate, and total reward. Our two experimental results in the training and testing phases revealed the baseline DRL with poor safety performance, while our proposed self-awareness attention-DQN can significantly improve the safety performance in intersection and roundabout scenarios.

\end{abstract}


\section{Introduction}

\IEEEPARstart{A}{utonomous} vehicle technologies for self-driving have attracted significant public attention through conveying huge benefits such as lower vehicle emissions, less traffic congestion, better fuel economy, or shorter travel time \cite{RN757, skeete2018level}. The current self-driving cars in the transportation system stay at simplified levels and need to consider efficacious and practical issues, including precise controls, costs, liability, privacy, and safety \cite{RN758, shneiderman2020bridging}.

Although traditional control theory and supervised learning applied to path planning of autonomous driving have been investigated since the last century \cite{RN760, bazzan2009opportunities, hussain2018autonomous}, reinforcement learning (RL) has recently demonstrated powerful performance, with a basis of enabling a vehicle to learn optimal control policies by interacting with the environment, without the need for prior knowledge or a large amount of training data \cite{wu2017flow}. Furthermore, deep reinforcement learning (DRL) models have achieved some notable successes \cite{RN676}. For example, AlphaGo \cite{RN674} defeated a human professional in the board game Go in 2015, which was the first time a computer Go program had defeated a professional human player in history. The deep Q network (DQN) \cite{RN673}, an off-policy method that adds a target network for stability and prioritised experience replay for efficient sampling, demonstrated high efficiency in playing 2600 Atari video games and outperformed humans just by learning from raw image pixels. Furthermore, advantage actor-critic (A2C) \cite{mnih2016asynchronous}, as another off-policy method that uses advantage estimates to calculate the value proposition for each action state pair, is a lightweight framework that uses the synchronised gradient to update that keeps the training more cohesive and potentially makes convergence faster. In addition, proximal policy optimisation (PPO) \cite{schulman2017proximal} is an on-policy method that provides better convergence with simplicity to implement and tune in most benchmark RL testing environments and was applied to traffic oscillations \cite{li2021reinforcement} and hierarchical guidance competitions \cite{cao2021reinforcement,cao2021weak}. These DRL research initiatives have stimulated more recent research areas on the application of DRL in autonomous driving and provided promising solutions in predictive perception, path and motion planning, driving strategies, and low-level controller design \cite {RN749}. However, they paid too much attention to evaluating the reward coverage returns and ignored the crucial safety factors in autonomous driving.

\section{Related Work}

The vehicle needs to reinforce learning from experience, usually at the cost of collisions, to gain autonomous knowledge and achieve higher rewards. Recent research on the DRL-based driving domain focused on the long-term accumulative reward or averaged reward as the critical performance metrics \cite{zhou2020smarts, RN655}. Nevertheless, the literature rarely paid attention to a safety measure to evaluate the autonomous performance of DRL models. One recent study \cite{RN775} placed a focal point on the vehicle's safety issue but only applied it to a straight four-lane highway road instead of a road traffic junction driving scenario.

To investigate the self-awareness mechanism for DRL, we paid heed to the self-attention module that was first introduced by \cite{RN654} to resolve the inability to translate complex sentences in natural language processing tasks, which enables the memorisation of long source sentences by introducing a dynamic context vector of importance weights. Afterwards, \cite{RN662} developed the transformer model, which is a multi-head self-attention module and achieved outstanding performance for translation tasks. Furthermore, \cite{RN726} used a variant form of the transformer model to accommodate the varying size of input data from the surrounding vehicles in autonomous driving, particularly in advance of addressing the limitation of the list of feature representations in a function approximation setting and enhancing the interpretability of the interaction patterns between traffic participants. However, the current explorations did not discover an attention mechanism to evaluate the safety concerns of an autonomous vehicle in the driving environment, specifically challenging road traffic junction scenarios.

In this work, we \textbf{aim} to (1) evaluate the safety concerns from four metrics (collision, success, freezing and reward) from three state-of-the-art baseline DRL models (DQN, A2C, and PPO) in autonomous driving, especially in complex road traffic junction driving environments, such as intersection and roundabout scenarios; (2) propose a self-awareness module for the DRL model to improve the safety performance in road traffic junction scenarios. The outcomes of this work will contribute to safe DRL for autonomous driving.

\section{Methodology}

\subsection{Baseline DRL Models}

\subsubsection{DQN}

DQN \cite{RN673} is a relatively typical DRL model inspired by \cite{RN744}, where the goal of the agent is to interact with the environment by selecting actions in a way that maximises cumulative future rewards. DQN approximates the optimal action-value function as follows:

\begin{equation} 
{Q^*(s,a)=\mathbb{E}_{s'\sim\epsilon}\left[r+\gamma\max_{a'}Q^*(s',a')|s,a\right]}
\end{equation}
where $Q^*(s, a)$ is the maximum sum of rewards discounted by $\gamma$ at each time step, achievable by following any strategy in a state $s$ and taking action $a$. This function follows the Bellman equation: if the optimal value $Q^*(s', a')$ of the state $s'$ at the next time step was known for all possible actions, then the optimal strategy is to select the action $a'$ maximising the expected value of $r+\gamma Q^*(s', a')$.

DQN has some advancements used for autonomous driving. First, DQN employs a mechanism called experience replay that stores the agent's experience $e_t=(s_t, a_t, r_t, s_{t+1})$ at each time step $t$ in a replay memory $D_t={e_1,\dotsc,e_t}$. During learning, samples of experience $(s, a, r, s')\sim U(D)$ are drawn uniformly at random from the replay memory, which removes the correlations in the observation sequence and smooths over changes in the data distribution. Second, DQN uses an iterative update to adjust the action values Q towards the target values. A neural network function approximator (Q-network) with weights $\theta$ is used to estimate the Q function. The Q-network updates only the target values periodically to reduce correlations with the target, and the Q-network is trained by minimising the following loss function $L$ at iteration $i$:

\begin{equation}
\newcommand*{\Scale}[2][4]{\scalebox{#1}{$#2$}}%
\Scale[0.9]{L_i(\theta_i)=
\mathbb{E}_{(s,a,r,s')\sim U(D)}\left[{\left(r+\gamma\max_{a'}Q(s',a';{\theta_i}^-)-Q(s,a;\theta_i)\right)}^2\right]}
\end{equation}
where $\theta_i$ denotes the parameters of the Q-network at iteration $i$. ${\theta_i}^-$ are the parameters of the target network at iteration $i$, which are held fixed between individual updates and are only updated with the Q-network parameters $\theta_i$ every certain step.

\subsubsection{A2C}
A2C is a synchronous, deterministic variant of asynchronous advantage actor-critic (A3C) \cite{mnih2016asynchronous}. For autonomous driving, A2C maintains a policy $\pi(a_t|s_t;\theta)$ and an estimate of the value function $V(s_t;\theta_v)$, which uses the same mix of $n$-step returns to update the policy and the value function. The policy and the value function are updated every $t_{max}$ actions or when a terminal state is reached. The update performed by A2C can be denoted as $\nabla_{\theta^{\prime}} \log \pi\left(a_{t} s_{t} ; \theta^{\prime}\right) A\left(s_{t}, a_{t} ; \theta, \theta_{v}\right)$, and $A(s_t,a_t;\theta,\theta_v)$ denotes the advantage function as follows:

\begin{equation}
\newcommand*{\Scale}[2][4]{\scalebox{#1}{$#2$}}%
\Scale[1]{A(s_t,a_t;\theta,\theta_v)=\sum_{i=0}^{k-1}\gamma^i r_{t+i}+\gamma^kV(s_{t+k};\theta_v)-V(s_t;\theta_v)}
\end{equation}
where $k$ denotes the number of actions taken since time step $t$ and is upper-bounded by $t_{max}$.

\subsubsection{PPO}
PPO is a model-free, actor-critic, and policy-gradient approach to maintain data efficiency and reliable performance \cite{schulman2017proximal}. In PPO, $\pi$ denotes the policy network optimised with its parameterisation $\theta$, and the policy network takes the state $s$ as the input and outputs an action $a$. PPO uses actor-critic architecture to enable learning of better policies by reformulating reward signals in terms of advantage $A$. The advantage function measures how good an action is compared to the other actions available in the state $s$. PPO maximises the surrogate objective function as follows:

\begin{equation}
\newcommand*{\Scale}[2][4]{\scalebox{#1}{$#2$}}%
\Scale[1]{L(\theta)=\hat{\mathbb{E}}_t[min({r_t(\theta)\hat{A}_t,clip(r_t(\theta),1-\epsilon,1+\epsilon)\hat{A}_t})]}
\end{equation}
where $L(\theta)$ is the policy gradient loss under parameterization $\theta$. $\hat{\mathbb{E}}_t$ denotes the empirically obtained expectation over a finite batch of samples, and $\hat{A}_t$ denotes an estimator of the advantage function at timestep $t$. $\epsilon$ is a hyperparameter for clipping, and $r_t(\theta)$ is the probability ratio formulised as:

\begin{equation}
r_t(\theta)=\frac{\pi_\theta(a_t,s_t)}{\pi_{\theta old}(a_t,s_t)}
\end{equation}

For autonomous driving, to increase sample efficiency, PPO uses importance sampling to obtain the expectation of samples gathered from an old policy $\pi_{\theta old}$ under the new policy $\pi_\theta$. As $\pi_\theta$ is refined, the two policies will diverge, and the estimation variance will increase. Therefore, the old policy is periodically updated to match the new policy. The clipping of the probability ratio to the range of $[1-\epsilon,1+\epsilon]$ ensures that the state transition function is similar between the two policies. The use of clipping discourages extensive policy updates that are outside of the comfortable zone.

\subsection{Self-Awareness Safety DRL}

The intuition of a self-awareness safety mechanism driven by the attention modules enables the ego vehicle to filter information and emphasises relevant vehicles that have potential conflicts with the ego vehicle's planned route, and the ego vehicle will be more informed to make decisions that could avoid collisions to stay safe. Following this motivation, we proposed a self-awareness safety DRL that employed the self-attention architecture \cite{RN726} and connected a normalisation layer to improve the training speed of multi-head attention based on the transformer framework \cite{RN765}. Second, we incorporated the proposed self-attention architecture with a baseline DRL model to build a self-awareness safety DRL, such as attention-DQN, to evaluate the safety performance of the ego vehicle in intersection and roundabout driving scenarios.

\subsubsection{Attention-DQN}

\begin{figure*}[!t]
\centering
\includegraphics[width=13cm]{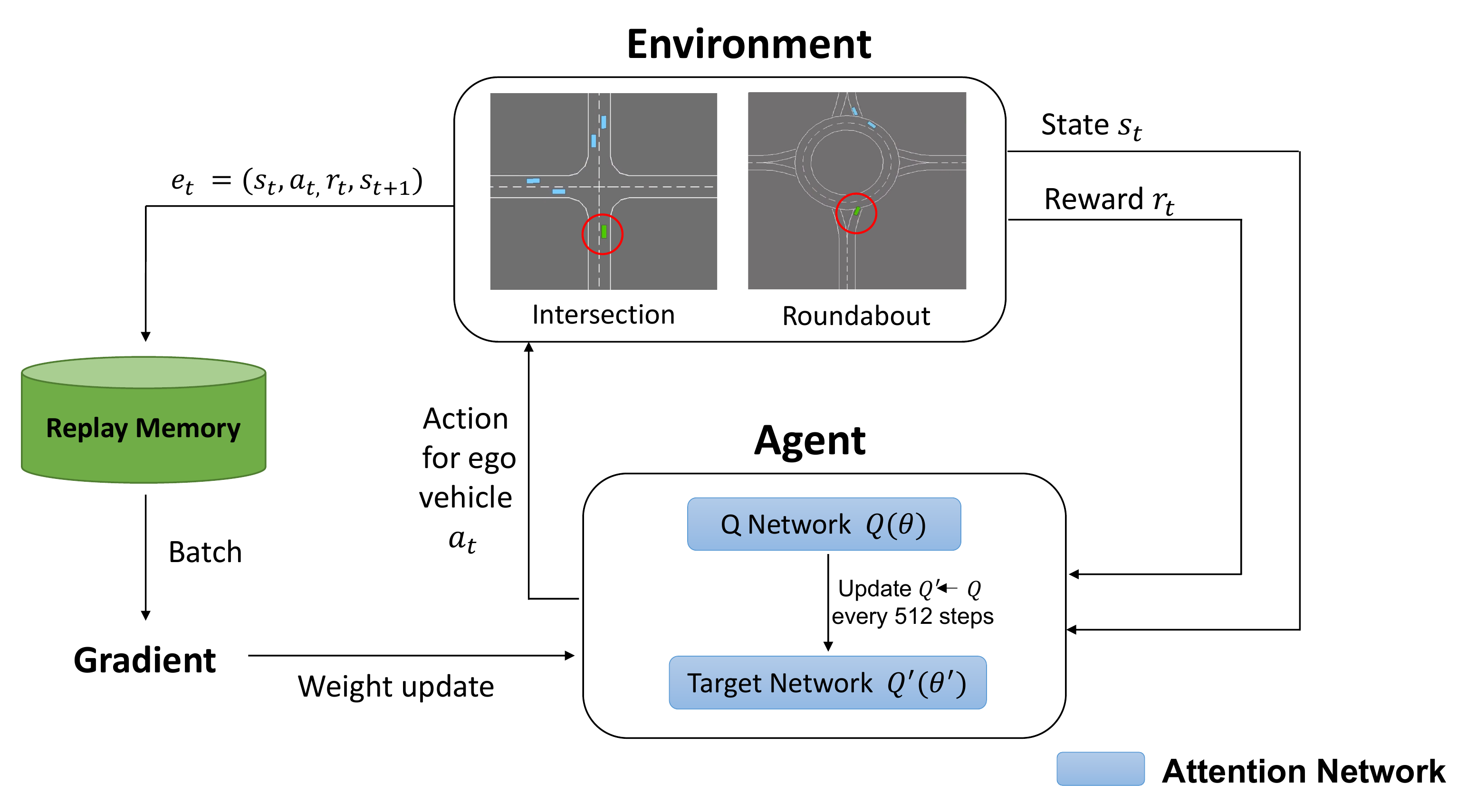}
\caption{Our proposed self-awareness safety DRL (attention-DQN). }
\label{fig_DQN_architecture}
\end{figure*}

As shown in \textbf{Figure \ref{fig_DQN_architecture}}, we presented a self-awareness safety DRL: attention-DQN, in a simulated road traffic junction environment including intersection and roundabout scenarios. The ego vehicle is an agent who observes the state $s_t$ of the environment and decides an action $a_t$; for instance, the vehicle accelerates forward. The environment then will give a reward $r_t$ as feedback and enter a new state $s_{t+1}$. A DQN agent contains two deep networks: the Q network and the target network; the Q network includes all updates during the training, while the target network is used to retrieve Q values. The target network parameters are only updated with the Q network parameters every 512 steps to reduce the correlations between Q-value updates and thus make the Q-network output's distribution stationary \cite{RN766}. DQN also uses the concept of experience replay to form a stable enough input dataset for training, i.e., all experiences in the form of $e_t=\left(s_t, a_t, r_t, s_{t+1}\right)$ are stored in the replay memory and sampled uniformly in a random manner.

\begin{figure}[!t]
\centering
\includegraphics[width=\columnwidth]{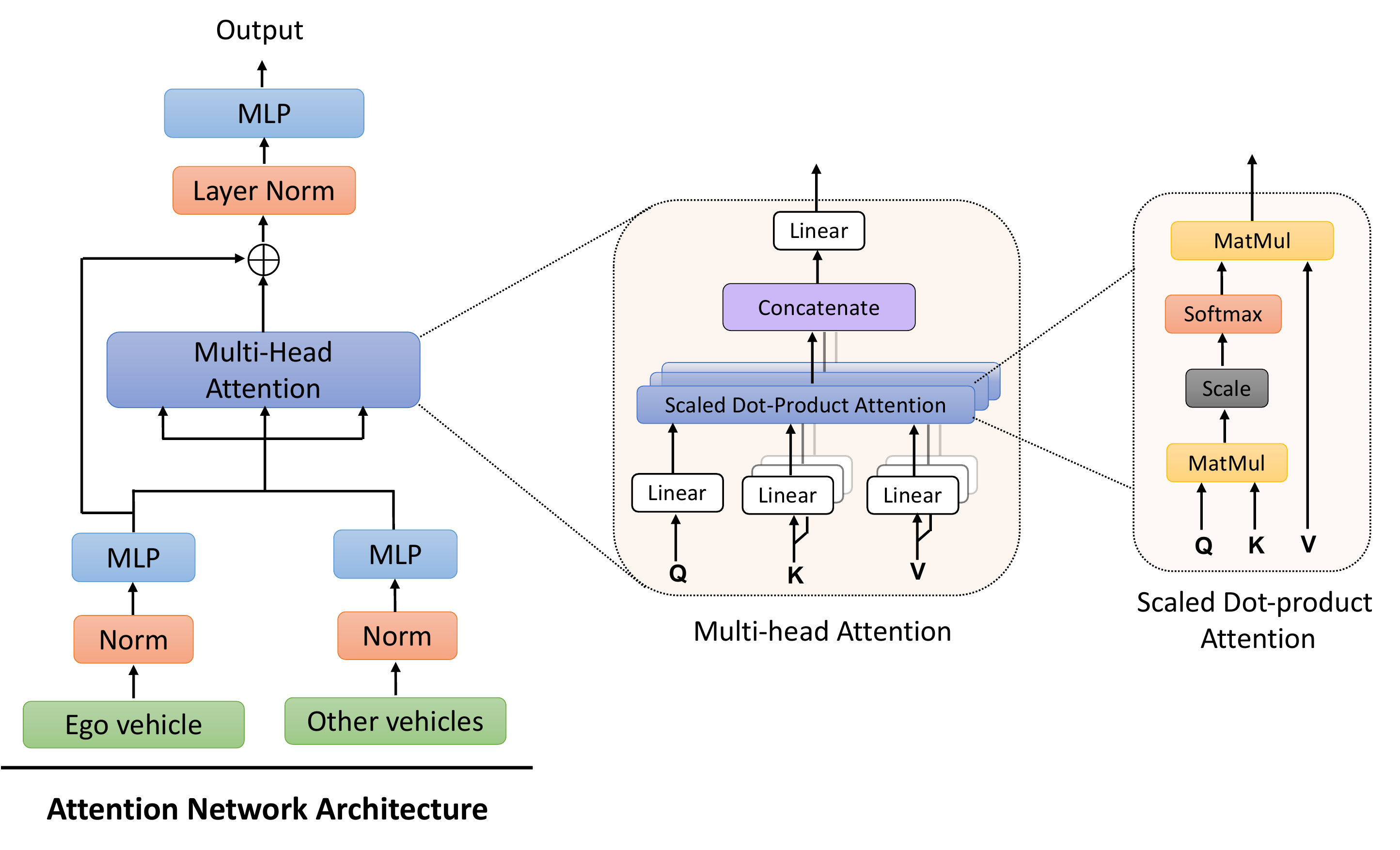}
\caption{The architecture of a self-attention network.}
\label{fig_architecture_attention_network}
\end{figure}

Furthermore, the two-deep networks in \textbf{Figure \ref{fig_DQN_architecture}} are considered attention networks, and the architecture of an attention network is illustrated in \textbf{Figure \ref{fig_architecture_attention_network}}. For an attention network, the observations containing vehicle features (i.e., location and velocity) are first split into the ego vehicle's inputs and those of other surrounding vehicles. The input observation information is normalised and then passed to a multilayer perceptron (MLP) network, and then the outputs from both MLPs are passed to a multi-head (i.e., two heads) attention block. Specifically, the multi-head attention block contains a single query $Q$ from the ego vehicle, keys of all the vehicles $K$ and values of all the vehicles $V$. Linear transformations are applied to the $Q$, $K$, and $V$, as well as the scale dot-product attention that is calculated from the transformed data. Then, the attention data from multiple heads are concatenated and linearly transformed before being passed to the next layer. Afterwards, the attention data from multiple heads are combined and added to the MLP output of the ego vehicle and then fed to a LayerNorm layer. Finally, the regularised data from the normalisation layer are passed to the output MLP to produce the action values \footnote {Source code is available at \href{https://github.com/czh513/DRL_Self-Awareness-Safety}{GitHub}. }.

\subsubsection{Default Single-query and Multi-query Attention-DQN}

\begin{figure}[!t]
\centering
\includegraphics[width=7cm]{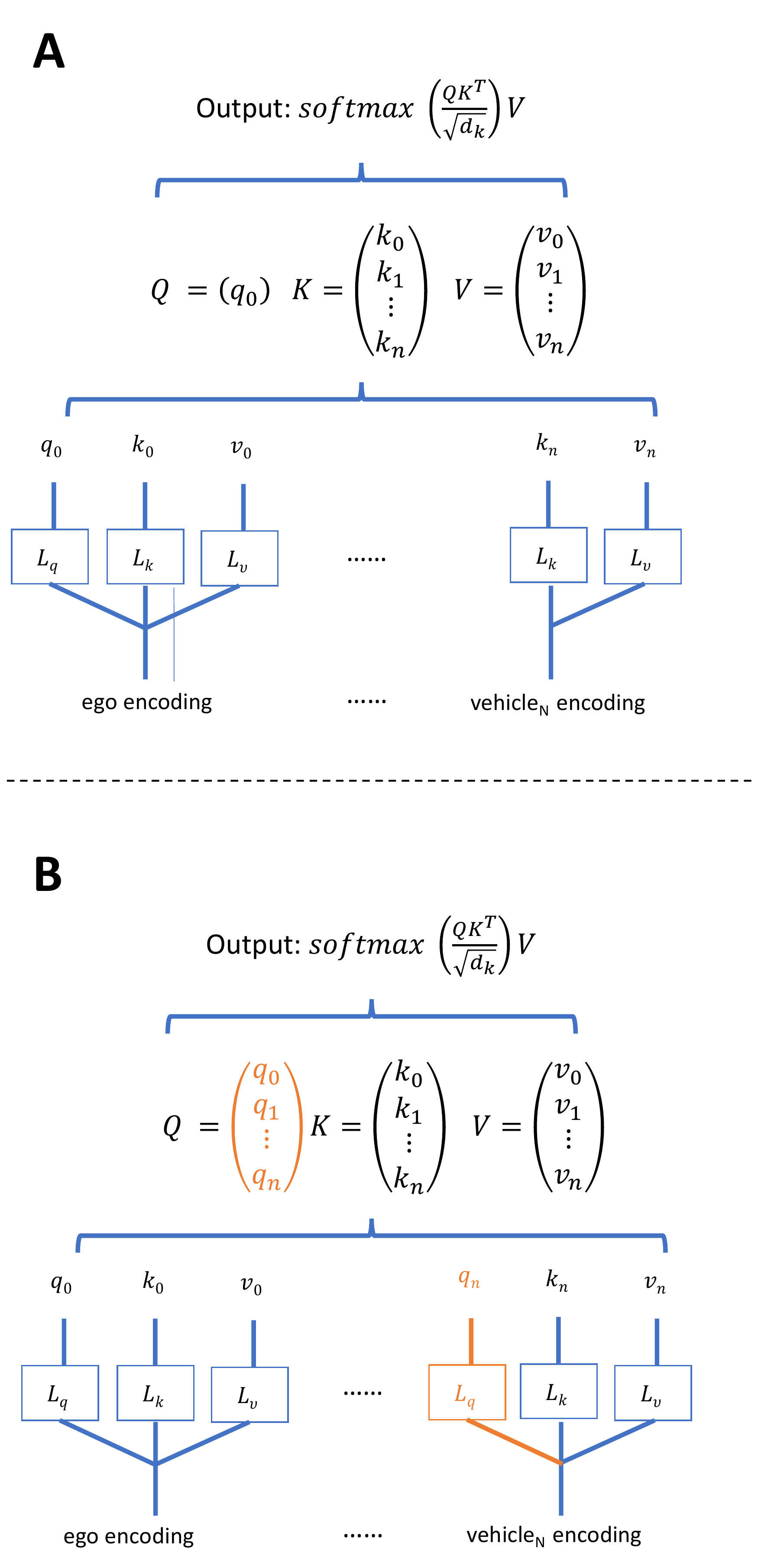}
\caption{The architecture of an attention head: (A) single-query attention head; (B) multi-query attention head.}
\label{fig_attention_head}
\end{figure}

Here, we introduced \textbf{Figure \ref{fig_attention_head}} to further present how an attention head works. In terms of the single-query attention architecture shown in \textbf{Figure \ref{fig_attention_head}-A}, first, the ego vehicle emits a default single query $Q= [q_0]$ to select a subset of vehicles according to the context. The query is computed with a linear projection $L_q$. Second, the query is compared to a set of keys $K= [k_0, …, k_N ]$ that contain descriptive features $k_i$ for each vehicle, and the keys are computed with a shared linear projection $L_k$. Third, the dot product of the query and keys $QK^T$ is computed to assess the similarities between them. The result is scaled by $( 1 / {\sqrt{d_k}})$, where $d_k$ is the dimension of keys, and a softmax function is applied to obtain the weights on the values $V= [v_0, …, v_N ]$. The values are also computed with a shared linear projection $L_v$. Please note that the keys $K$ and values $V$ are concatenated from all vehicles, whereas the query $Q$ is only produced by the ego vehicle. The output attention matrix is formalised as:

\begin{equation}
\label{eqn_attention}
output=softmax\left(\frac{QK^T}{\sqrt{d_k}}\right)V
\end{equation}

In addition, we explored a multi-query attention architecture, marked as "MultiQ-attention-DQN", as demonstrated in \textbf{Figure \ref{fig_attention_head}-B}, which tweaked from the single-query attention architecture to observe the performance variations, and the main differences were highlighted in orange. Here, we include queries from all the vehicles instead of using the default single query from the ego vehicle.

\section{Experiments}

\subsection{Driving Environment}

\begin{figure}[!t]
\centering
\includegraphics[width=\columnwidth]{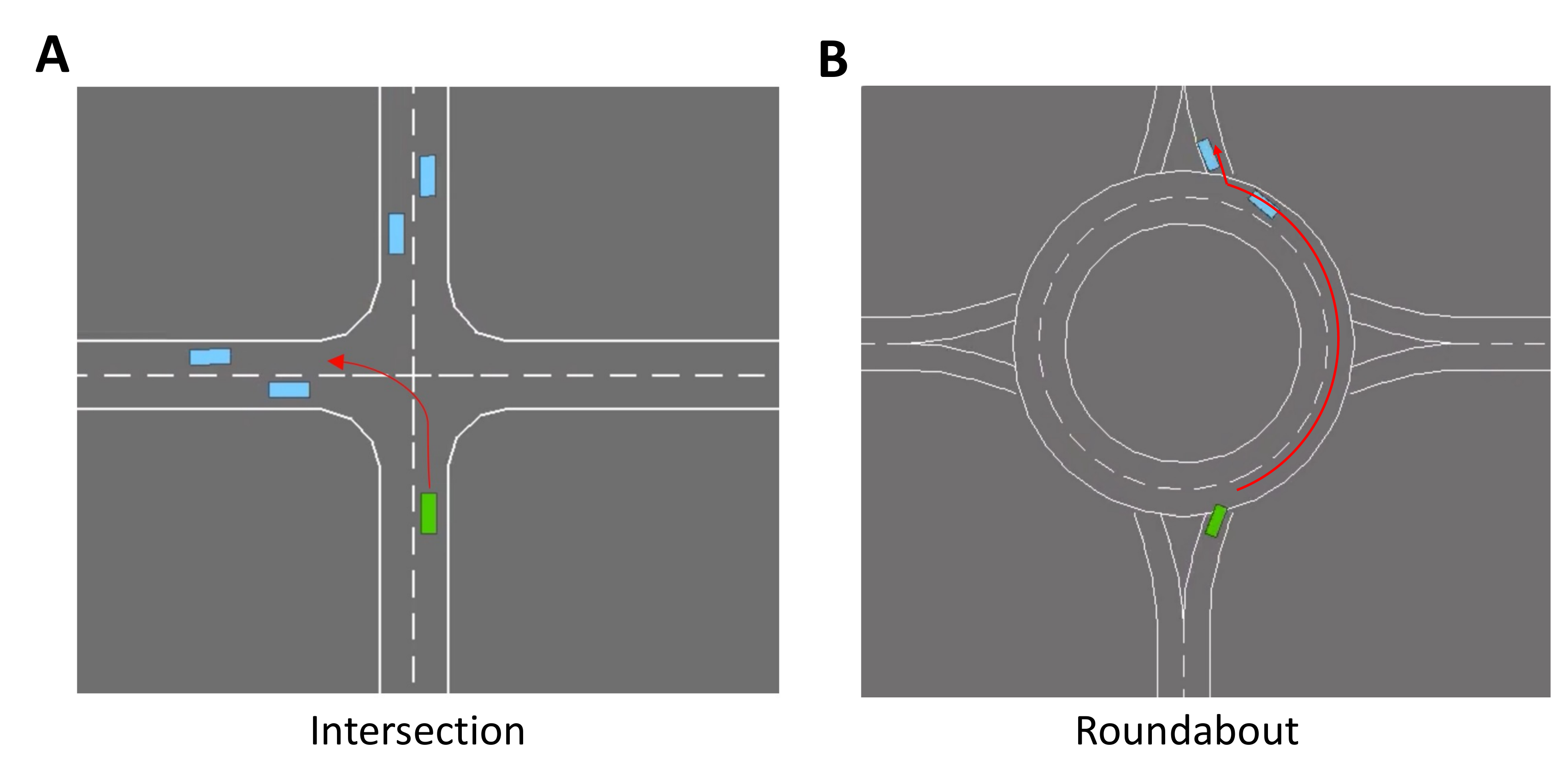}
\caption{Two representative driving scenarios in the road traffic junctions: (A) intersection; (B) roundabout. The ego vehicle with green colour is driving as guided by the red arrow to reach the goal.}
\label{fig_scenario_eg}
\end{figure}

In this study, we used two representative driving scenarios, intersection and roundabout, as shown in \textbf{Figure \ref{fig_scenario_eg}}, to investigate the safety of DRL for road traffic junction environments based on the highway simulation system \cite{RN754}. We suggested intersection (\textbf{Figure \ref{fig_scenario_eg}-A}) and roundabout (\textbf{Figure \ref{fig_scenario_eg}-B}) scenarios characterised as relatively challenging driving interflow environments. It includes one ego vehicle trying to cross the traffic and arrive at a destination, and several surrounding vehicles spawned randomly. The ego vehicle needs to make decisions such as turning left, changing lanes, accelerating and decelerating while avoiding colliding with other vehicles.

\subsubsection{Intersection Scenario}

\begin{itemize}[leftmargin=*]

\item \textit{Settings}. The intersection scene comprises two roads crossing perpendicularly and stretching towards four directions (north, south, east, and west). The roads are populated with the ego vehicle and several surrounding vehicles. The positions, velocities and destinations of the other vehicles are randomly initialised. The ego vehicle drives from 60 metres south to the intersection, and the task of the ego vehicle is to cross the intersection and turn left (west). The goal will be achieved (i.e., destination arrived) if the ego vehicle could turn left at the intersection and drive 25 metres or more from the turning point in 13 seconds. Please note that the horizontal road has the right of way in the intersection scenario as for the road priority. \textbf{Figure \ref{fig_scenario_eg}-A} is a showcase of the intersection scenario where the goal of the ego vehicle with green colour is to turn left as guided by the red arrow.

\item \textit{Observations}. The vehicle observations of the intersection scenario follow the kinematic bicycle model \cite{RN755}, which implements a simplistic behaviour and considers the same lane interactions to prevent potential lateral collisions \cite{RN754}: each vehicle predicts future positions of neighbouring vehicles over a three-second horizon. If a collision with a neighbour is predicted, the yielding vehicle is decided according to the road priorities. The chosen yielding vehicle will brake until the collision prediction clears.

\item \textit{Actions}. The ego vehicle in the intersection scenario is designed to operate by selecting one from a finite set of actions \textit{A = \{slower, no-operation, faster\}}, where the vehicle can choose to slow down, keep the constant speed and accelerate.

\item \textit{Rewards}. The reward of the ego vehicle for the intersection scenario is designed as follows: the ego vehicle receives a reward of 1 if it drives at the maximum speed or if it arrives at the destination and receives a reward of -5 if a collision happens. This reward design encourages the ego vehicle to cross the intersection and arrive at the destination as soon as possible while avoiding collisions simultaneously.

\end{itemize}

\subsubsection{Roundabout Scenario}

\begin{itemize}[leftmargin=*]

\item \textit{Settings}. The scene of the roundabout scenario is composed of a two-lane roundabout with four entrances/exits (north, south, east, and west), and the roads are occupied with the ego vehicle and several surrounding vehicles. The positions, velocities and destinations of the other vehicles are initialised randomly. The ego vehicle drives from 125 metres south to the roundabout, and the task of the ego vehicle is to cross the roundabout and take the second exit (north). The goal can be accomplished if the ego vehicle can successfully take the second exit at the roundabout and drive 10 metres or more from the exit point in 13 seconds. Please note that vehicles in the roundabout have the right of way as for the road priority. As presented in \textbf{Figure \ref{fig_scenario_eg}-B} for a showcase of the roundabout scenario, the goal of the ego vehicle is to exit the roundabout following the red arrow.

\item \textit{Observations}. The vehicle observations of the roundabout scenario are applied to the kinematics type as well. The ego vehicle observes and learns from a $V*F$ array, where $V$ is the number of nearby vehicles, and $F$ is the size of features observed. In our setting, the ego vehicle observes a feature set of size 7: $\{p,x,y,v_x,v_y,cos_h,sin_h\}$, where $p$ represents whether a vehicle is in the row (whether a vehicle was observed), $x$ and $y$ describe the location of the vehicle, $v_x$ and $v_y$ denote the $x$ and $y$ axes of velocity, and $cos_h$ and $sin_h$ describe the heading direction of the vehicle.

\item \textit{Actions}. The actions of the ego vehicle in the roundabout scenario are selected from a finite set: \textit{A = \{lane\_{left}, lane\_{right}, idle, faster, slower\}}, implying that the vehicle can choose to change to the left/right lane, keep the same speed and lane, and accelerate and decelerate.

\item \textit{Rewards}. The reward of the ego vehicle for the roundabout scenario is arranged as follows: the ego vehicle receives a reward of 0.5 if it drives at the maximum speed and receives a reward of -1 if a collision happens. Every time the vehicle changes a lane, it will be awarded as -0.05. We also designed a success reward of 1 if the ego vehicle arrives at the destination, encouraging the ego vehicle to reach the destination. We suppose that this reward design encourages the vehicle to cross the roundabout and arrive at the destination as soon as possible while avoiding collisions or unnecessary lane changing simultaneously.

\end{itemize}

\subsection{Evaluation Metrics}

Our study defined four evaluation metrics heavily considering safety concerns (collision, success, and freezing) and the standard encouraging total reward to assess three baseline DRL models (DQN, PPO, and A2C) and our proposed self-awareness safety DRL: attention-DQN.

\begin{itemize}[leftmargin=*]

\item \textit{Collision rate}. It calculates the percentage of the episodes where a collision happens to the ego vehicle, reflecting the safety level of the DRL models used to train and test the performance of an autonomous vehicle.

\item \textit{Success rate}. It counts the percentage of the episodes where the ego vehicle successfully reaches the destination within the allowed duration. It also considers the functionality of the autonomous vehicle, which evaluates the ability of a DRL model to learn behaviour to complete the desired task. In our experiment, the driving scenario has a timeout setting representing the maximum duration to cross the intersection. Vehicles exceeding this duration are examined as unacceptably slow, which aligns with real-world driving scenarios where cars moving extremely slowly on the road are usually unacceptable.

\item \textit{Freezing rate}. It determines the percentage of episodes where the ego vehicle freezes or moves extremely slowly in the driving environment. In the transportation system, it is common to observe a freezing robot problem, where the ego vehicle acts overcautiously and freezes or moves extremely slowly on the road. In this case, the ego vehicle can be a serious safety hazard to traffic, such as the case of causing serial rear-end collisions. This metric is vital for straightforward performance comparisons to evaluate whether our proposed module can help alleviate the freezing robot problem, since the frozen ego vehicle will neither contribute to collision rate nor success rate. In our experiment, the freezing rate is calculated by the residual value from the success rate and collision rate as follows:

\begin{equation}
    Rate_{freezing}=100\%-Rate_{collision}-Rate_{success}
    \end{equation}

\item \textit{Total reward}. It is a typical metric commonly used to evaluate ego vehicle premium performance over several episodes. Our experiment collected the total reward that the ego vehicle receives after taking a set of actions in each episode. As the ego vehicle learns to drive better, the total reward the ego vehicle acquires in each episode would increase with more learning episodes.

\end{itemize}

\subsection{Experimental Procedure}

We conducted two experiments in this study: in experiment 1, three baseline DRL models, DQN, A2C, and PPO, were selected to train the ego vehicle and then evaluate the safety performance based on four defined metrics in the intersection and roundabout driving scenarios; in experiment 2, we trained the ego vehicle in the intersection and roundabout driving scenarios using the proposed single-query and multi-query attention-DQN to compare the performance with the baseline DQN model. In addition, we presented the safety performance of the above DRL models from two experiments during the training and testing phases.

\subsubsection{Experiment 1}

During the training phase, the ego vehicle was drilled by three baseline DRL models (DQN, A2C, and PPO) in intersection and roundabout driving scenarios. We observed that the collision rate and success rate in each DRL model converged after 60,000 training episodes with 5 runs. In the testing phase, we evaluated the trained DRL models over 100 episodes for 100 trials to collect the testing data and tested each for 20 trials. For each trial, the number of collisions and successes during the 100 testing episodes were recorded to represent the collision rate and success rate. Additionally, the freezing rate was calculated through the residual value from the collected collision rate and success rate.

\subsubsection{Experiment 2}

We trained the ego vehicle using default single-query and multi-query attention-DQN in intersection and roundabout driving scenarios for 60,000 episodes with 5 runs. Please note that the hyperparameters we used for attention-based DQN are the same as the baseline DQN for a fair comparison. In the testing phase, we evaluated the trained models from single-query and multi-query attention-DQN over 100 episodes for 100 trials to collect the testing data and tested each for 20 trials. We recorded the number of collisions and the number of successes during the 100 episodes to indicate the collision rate and success rate for each trial. The freezing rate is the calculated residual value from the collision rate and the success rate.
\section{Results}

\subsection{Evaluation of Experiment 1}

\subsubsection{Training Phase}

\begin{itemize}[leftmargin=*]

\item \textit{Intersection scenario}. As shown in \textbf{Figure \ref{fig_exp1_intersection}}, we presented the training evolution performance of collision rate (\textbf{Figure \ref{fig_exp1_intersection}-A}), success rate (\textbf{Figure \ref{fig_exp1_intersection}-B}), freezing rate (\textbf{Figure \ref{fig_exp1_intersection}-C}) and total reward (\textbf{Figure \ref{fig_exp1_intersection}-D}) of the ego vehicle in the intersection scenario, trained by three baseline DRL models (DQN, PPO, and A2C) over 60,000 episodes and repeated 5 times, and the displayed values were averaged with 95\% confidence interval. We found that the collision rate of the ego vehicles trained by three baseline DRL models reached above 60\% at the end of the training steps, particularly in DQN, which revealed the lowest collision rate during the training process. We also noticed that the success rates of PPO and A2C reached approximately the same value at 35\%, and the success rate of DQN reached approximately 28\%, which was reflected in the total reward curves. In addition, the freezing rates of the ego vehicles trained by PPO and A2C reached approximately 0, starting from 15,000 episodes, whereas the freezing rate of DQN decreased to approximately 8\%, being the best performance among all baseline DRL models.

\begin{figure}[!t]
\centering
\includegraphics[width=\columnwidth]{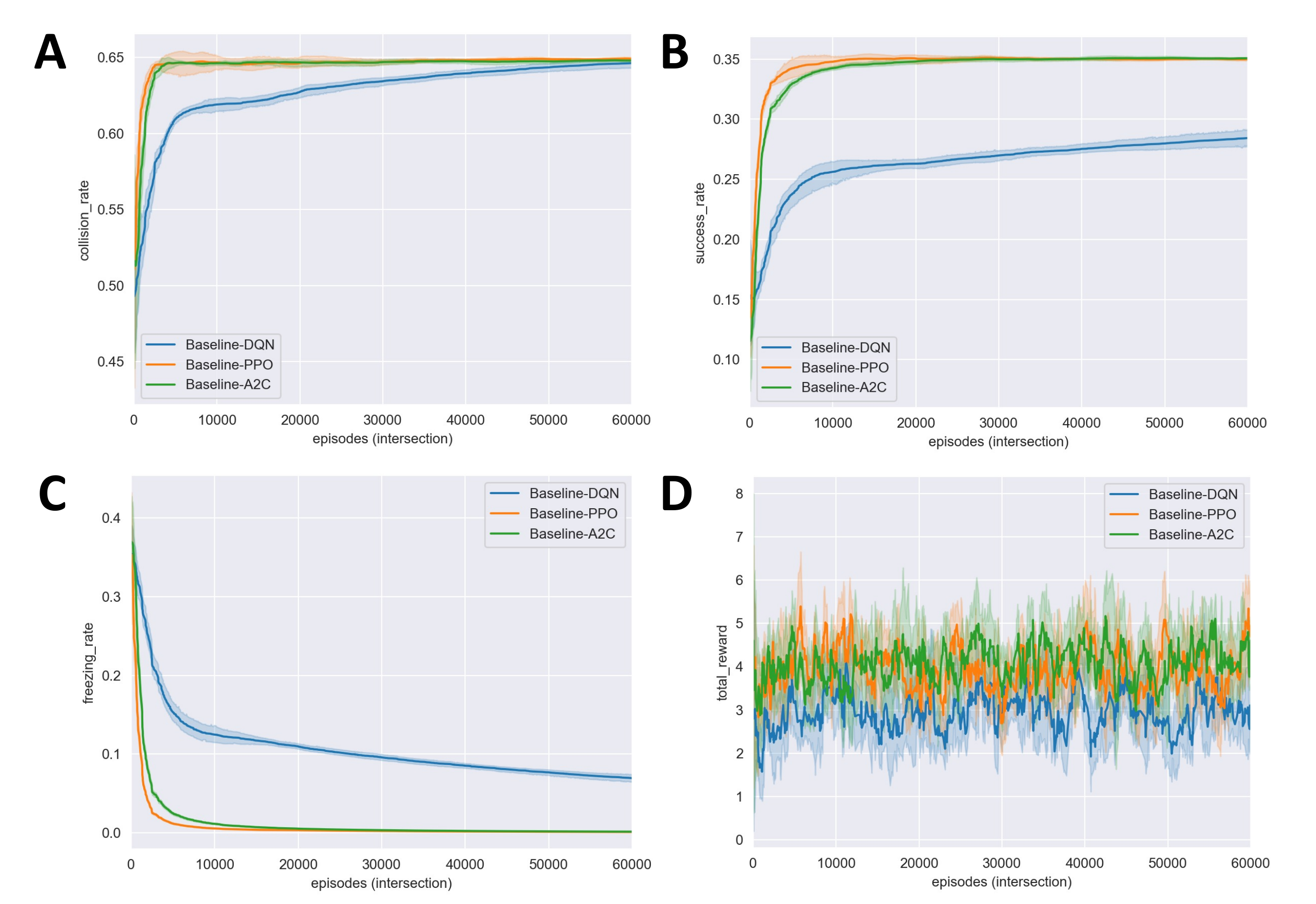}
\caption{Intersection scenario: the training performance of the ego vechile with four metrics: (A) collision rate, (B) success rate, (C) freezing rate, and (D) total reward using the baseline DRL models (DQN, A2C, and PPO).}
\label{fig_exp1_intersection}
\end{figure}

\item \textit{Roundabout scenario}. As presented in \textbf{Figure \ref{fig_exp1_roundabout}}, we demonstrated the training evolution of the performance of the collision rate (\textbf{Figure \ref{fig_exp1_roundabout}-A}), success rate (\textbf{Figure \ref{fig_exp1_roundabout}-B}), freezing rate (\textbf{Figure \ref{fig_exp1_roundabout}-C}) and total reward (\textbf{Figure \ref{fig_exp1_roundabout}-D}) of the ego vehicle during training in the roundabout scenario, trained by three baseline DRL models (DQN, PPO, and A2C) over 60,000 episodes and repeated 5 times, and the displayed values were averaged with a 95\% confidence interval. We recognised that the collision rate of the ego vehicles trained by all baseline DRL models was below 25\% at the end of the training, in particular A2C being visible as the lowest at approximately 5\%. In contrast, the freezing rates of the ego vehicles trained by A2C and DQN both reached above 60\%, specific to A2C being approximately 73\% and DQN being approximately 63\%. In addition, PPO revealed the lowest freezing rate and the highest success rate in baseline DRL models.

\begin{figure}[!t]
\centering
\includegraphics[width=\columnwidth]{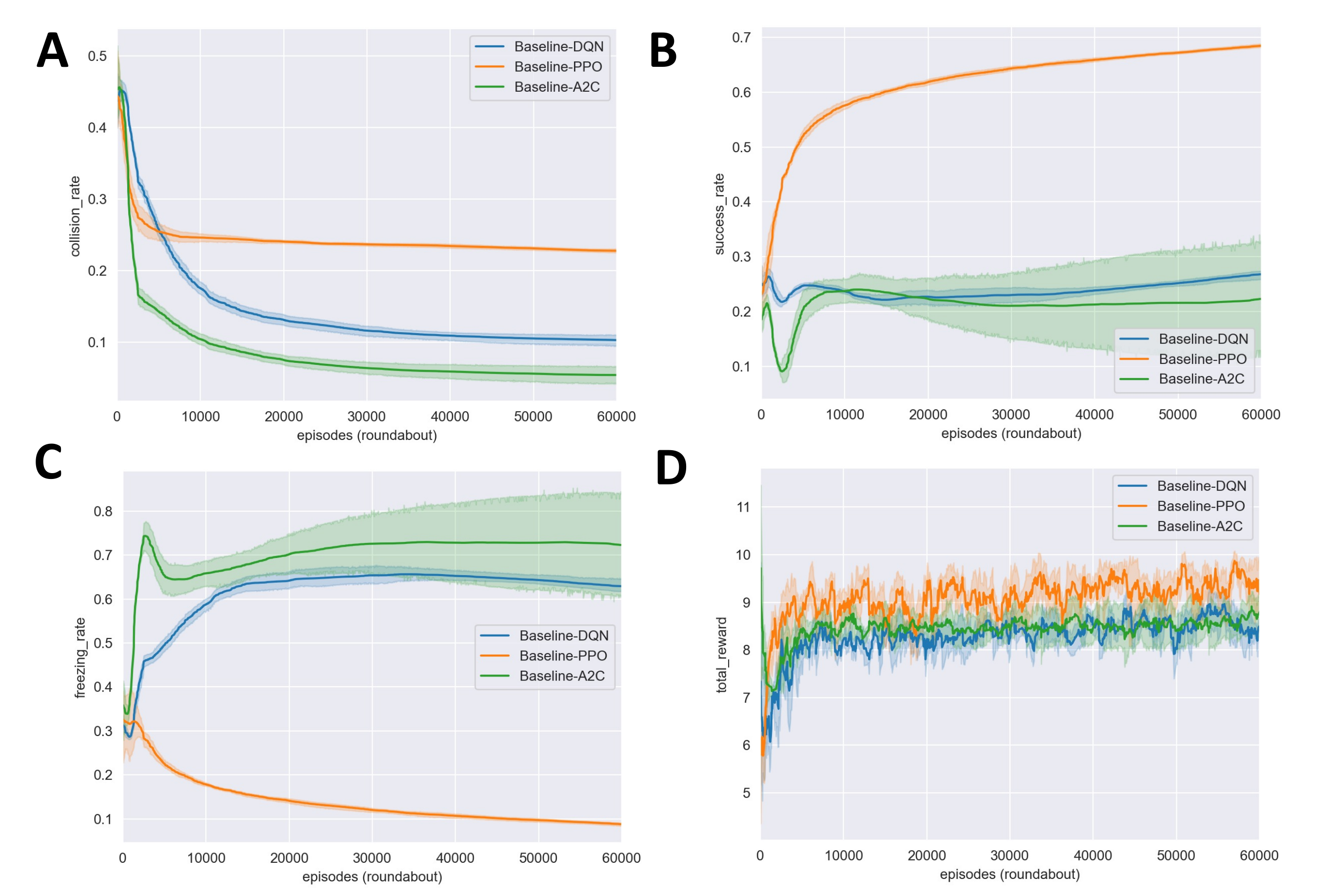}
\caption{Roundabout scenario: the training performance of the ego vechile with four metrics: (A) collision rate, (B) success rate, (C) freezing rate, and (D) total reward using the baseline DRL models (DQN, A2C, and PPO).}
\label{fig_exp1_roundabout}
\end{figure}

\end{itemize}

\subsubsection{Testing Phase}

\textbf{Table \ref{table_1}} shows the collision rates, success rates, and freezing rates of the ego vehicles tested in the intersection and roundabout scenario using three baseline DRL models over 100 episodes and averaged across 100 runs (10,000 testing episodes in total for each model). Regarding the intersection scenario, the collision rates of the ego vehicles trained by all three baseline models were above 50\%, with PPO being the highest at 64.45\% and DQN being the lowest at 56.88\%. In terms of success rate, DQN also came the lowest value, at 29.23\%. This was caused by DQN being the only case that experienced a freezing robot problem, with a freezing rate of 13.89\%. The PPO and A2C models did not have freezing vehicles during the testing phase. Moving to the testing phase of the roundabout scenario, A2C demonstrated the lowest collision rate of 7.2\%, followed by DQN of 14.7\% and PPO of 20.47\%. DQN had the lowest success rate and the highest freezing rate, whereas PPO had the highest success rate and lowest freezing rate, with A2C being the second in both metrics.

\begin{table}[!ht]
\caption {The testing performance of three baseline DRL models in intersection and roundabout scenarios.} \label{table_1} 
\begin{adjustbox}{width=\columnwidth,center}
\begin{tabular}{@{}ccccc@{}}
\toprule
\textbf{Scenarios}                                           & \textbf{\begin{tabular}[c]{@{}c@{}}Metrics\\ mean (std) \%\end{tabular}} & \textbf{DQN}  & \textbf{A2C} & \textbf{PPO} \\ \midrule
\multicolumn{1}{c|}{\multirow{3}{*}{\textbf{Intersection}}} & \textit{Collision Rate}                                             & 56.88 (5.39)  & 65.51 (5.08) & 64.45 (4.93) \\
\multicolumn{1}{c|}{}                                       & \textit{Success Rate}                                               & 29.23 (4.99)  & 34.49 (5.08) & 35.54 (4.93) \\
\multicolumn{1}{c|}{}                                       & \textit{Freezing Rate}                                              & 13.89 (3.81)  & 0 (0)        & 0 (0)        \\ \midrule
\multicolumn{1}{c|}{\multirow{3}{*}{\textbf{Roundabout}}}   & \textit{Collision Rate}                                             & 14.7 (7.24)   & 7.2 (3.03)   & 20.47 (4.78) \\
\multicolumn{1}{c|}{}                                       & \textit{Success Rate}                                               & 33.99 (5.66)  & 46.35 (7.70) & 74.42 (4.91) \\
\multicolumn{1}{c|}{}                                       & \textit{Freezing Rate}                                              & 51.31 (10.33) & 46.45 (8.58) & 5.11 (5.39)  \\ \cmidrule(l){2-5} 
\end{tabular}
\end{adjustbox}
\end{table}

\subsubsection{Our Findings}

We observed significant room to improve safety performance in the testing phase, as shown in \textbf{Table \ref{table_1}}, especially the collision rate of the DRL models in DQN. In the intersection scenario, we observed a high collision rate - more than 50\% from all three baseline DRL models (56.88\% for DQN, 65.51\% for A2C, and 64.45\% for PPO), suggesting that the autonomous vehicle in the road traffic junction driving environment is far from safe. Furthermore, we noticed that the collision rate in the roundabout scenario is lower (<20\%) than that in the intersection scenario, but it is mainly caused by the high freezing rate that does not indeed contribute to safety concerns. Specifically, the freezing robot problem observed in the roundabout scenario was remarked during the training phase of all baseline DRL models, especially to DQN, which posed more than half of the freezing episodes (51.31\%). Considering the above performance results, we chose DQN to investigate further whether the proposed self-attention module could decrease the collision rate and alleviate the freezing robot problem in intersection and roundabout driving scenarios.

\begin{figure}[!t]
\centering
\includegraphics[width=\columnwidth]{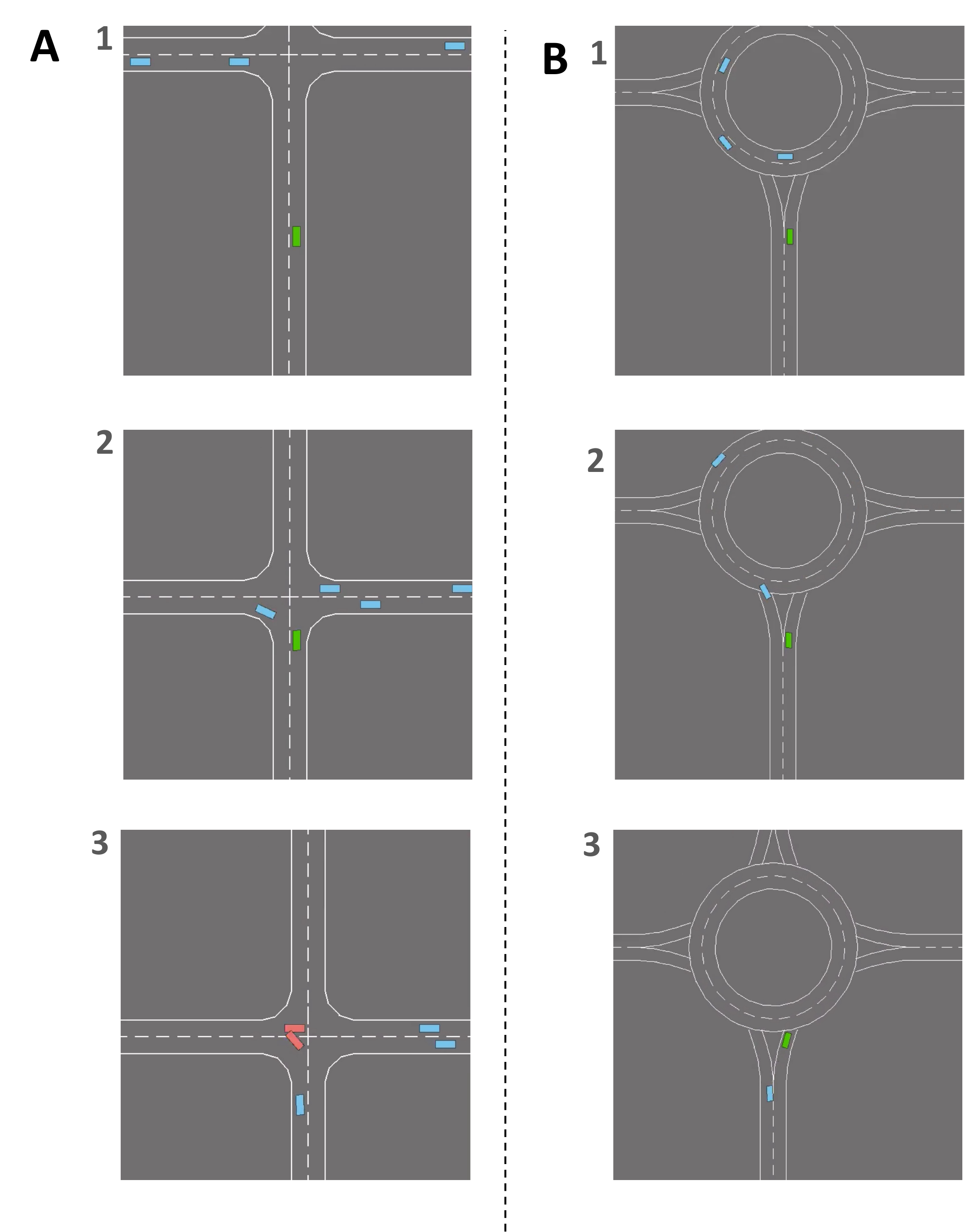}
\caption{A showcase of testing episodes using DQN in (A) intersection and (B) roundabout scenarios.}
\label{fig_exp1_findings}
\end{figure}

In addition, we demonstrated a showcase in \textbf{Figure \ref{fig_exp1_findings}} about testing episodes using DQN in intersection and roundabout scenarios to further clarify our findings. The goal of the ego vehicle is to turn left in the intersection scenario (\textbf{Figure \ref{fig_exp1_findings}-A}), and the training behaviours of the ego vehicle follows: (1) the ego vehicle began from the starting point, driving towards the intersection; (2) the ego vehicle reached the intersection entry, and some other vehicles came from left and right. The surrounding vehicle on the left probably would not collide with the ego vehicle since it was turning right, but the surrounding vehicle from the right might collide with the ego vehicle; (3) the ego vehicle made the decision to turn left and collided with the incoming car from the right. For the roundabout scenario (\textbf{Figure \ref{fig_exp1_findings}-B}), the goal of the ego vehicle is to take the north (second) exit from the roundabout, and the training behaviours of the ego vehicle follow as follows: (1) the ego vehicle began from the starting point, driving towards the roundabout, and there were other vehicles in the roundabout that might collide with the ego vehicle; (2) the ego vehicle stayed basically where it was when it seemed safe to enter the roundabout; (3) the roundabout was completely clear of vehicles, and the ego vehicle finally moved forward before time was out, and the episode ended.

\subsection{Evaluation of Experiment 2}

\subsubsection{Training Phase}

\begin{itemize}[leftmargin=*]

\item \textit{Intersection scenario}. As shown in \textbf{Figure \ref{fig_exp2_intersection}}, we presented the training evolution of collision rate (\textbf{Figure \ref{fig_exp2_intersection}-A}), success rate (\textbf{Figure \ref{fig_exp2_intersection}-B}), freezing rate (\textbf{Figure \ref{fig_exp2_intersection}-C}) and total reward (\textbf{Figure \ref{fig_exp2_intersection}-D}) of the ego vehicle in the intersection scenario, learned by DQN, default (single-query) and multiQ (multi-query) attention-DQN over 60,000 episodes, which displayed by the mean values averaged from 5 runs, with 95\% confidence interval. We observed that the ego vehicle trained by the default attention-DQN (the orange colour) achieved the lowest collision rate and the highest success rate, which was demonstrated by a significant drop in the collision rate, from approximately 67\% of the DQN to approximately 20\% of the attention-DQN, as well as by a considerable increase from 28\% of DQN to 49\% of the attention-DQN. The changes in collision and success rates were also reflected in the total reward. We also noticed that the multiQ-attention-DQN (the green colour) exhibited a similar performance yet not as strong compared with default attention-DQN.

\begin{figure}[!t]
\centering
\includegraphics[width=\columnwidth]{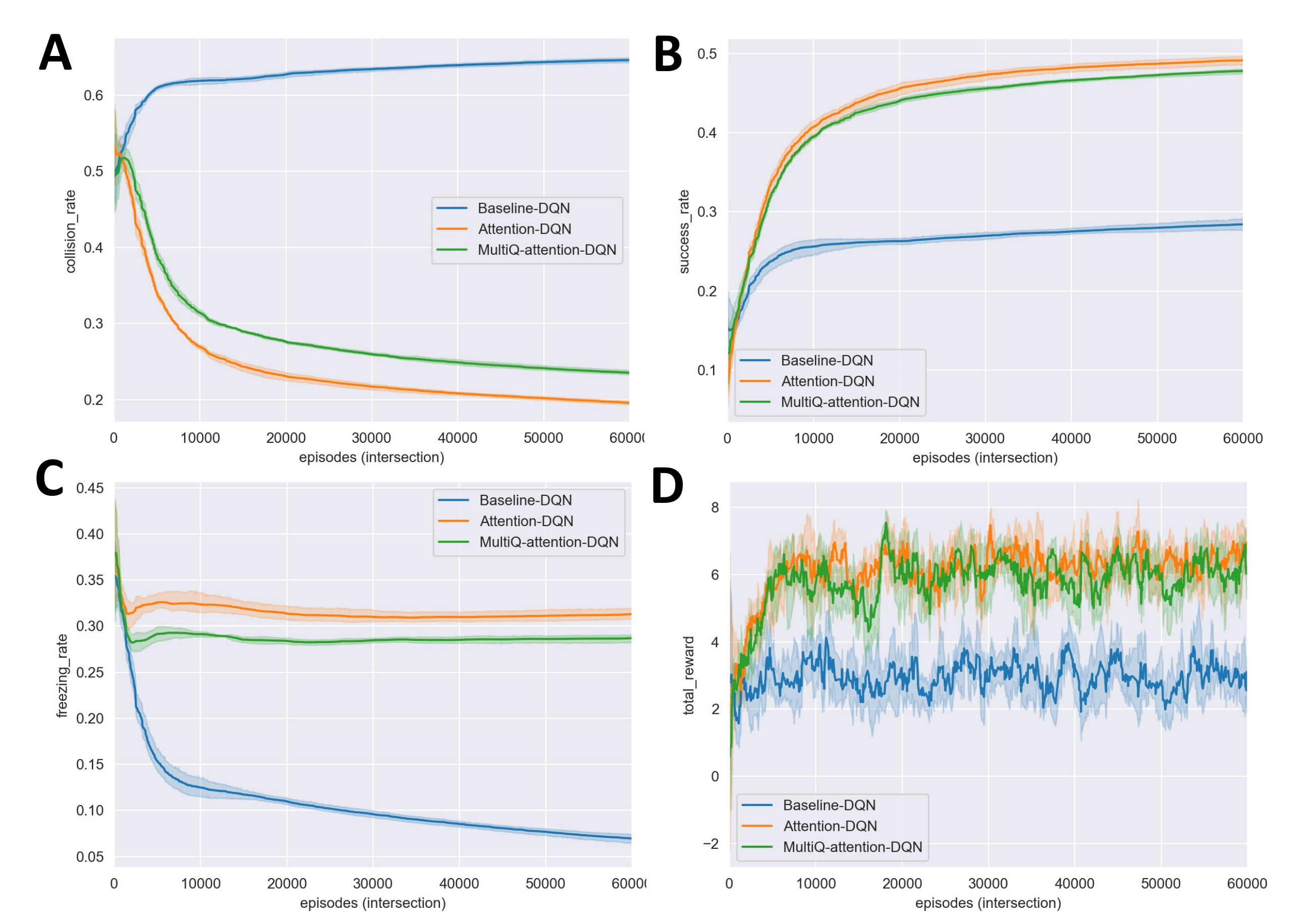}
\caption{Intersection scenario: the training performance of the ego vechile with four metrics: (A) collision rate, (B) success rate, (C) freezing rate, and (D) total reward using DQN, single-query and multi-query attention-DQN.}
\label{fig_exp2_intersection}
\end{figure}

\item \textit{Roundabout scenario}. \textbf{Figure \ref{fig_exp2_roundabout}} demonstrates the training evolution of the collision rate (\textbf{Figure \ref{fig_exp2_roundabout}-A}), success rate (\textbf{Figure \ref{fig_exp2_roundabout}-B}), freezing rate (\textbf{Figure \ref{fig_exp2_roundabout}-C}) and total reward (\textbf{Figure \ref{fig_exp2_roundabout}-D}) of the ego vehicles in the roundabout scenario learned by DQN, default (single-query) and multiQ (multi-query) attention-DQN over 60,000 episodes, which are displayed by the mean values averaged from 5 runs, with a 95\% confidence interval. We observed that the ego vehicle trained by attention-DQN was quickest to avoid collisions during the beginning training period, such as the curve's steepness between 1 to 10,000 episodes, and then reached the low collision rate of 12\%at episode 60,000 finally. For the success rate, the ego vehicle trained by the attention-DQN achieved the highest performance by presenting approximately 74\%, which is a significant improvement compared to 27\% introduced by DQN. In addition, the freezing rate dropped from 63\% of DQN to approximately 15\% of attention-DQN.

\begin{figure}[!t]
\centering
\includegraphics[width=\columnwidth]{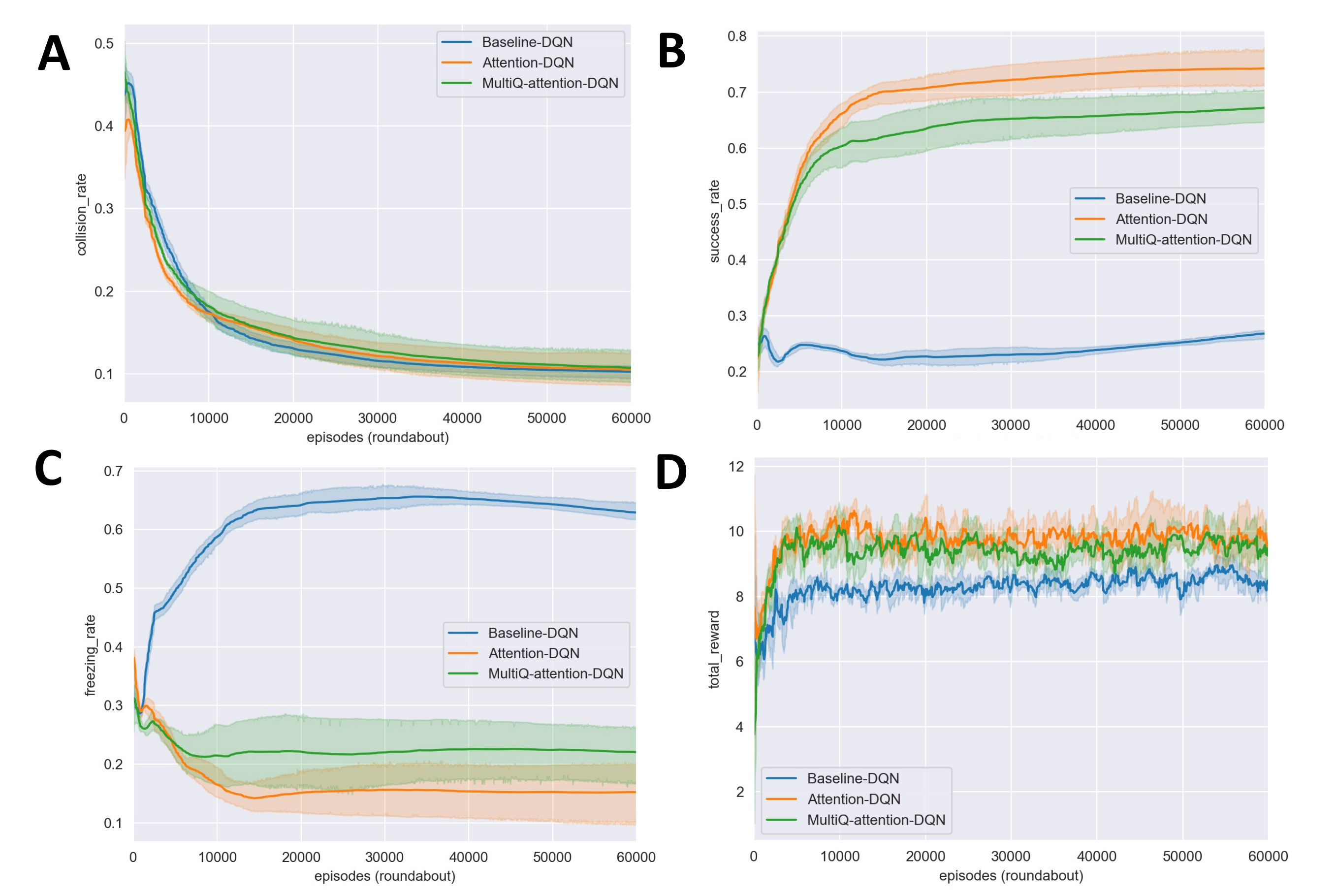}
\caption{Roundabout scenario: the training performance of the ego vechile with four metrics: (A) collision rate, (B) success rate, (C) freezing rate, and (D) total reward using DQN, single-query and multi-query attention-DQN.}
\label{fig_exp2_roundabout}
\end{figure}

\end{itemize}

\subsubsection{Testing Phase}

As shown in \textbf{Table \ref{table_2}}, we summarised the collision rate, success rate, and freezing rate of the ego vehicle tested by trained DRL models (DQN, default attention-DQN and MultiQ-attention-DQN) over 100 episodes and averaged across 100 runs (10,000 testing episodes in total for each DRL model) in the intersection and roundabout scenarios. In the intersection scenario, the testing performance of the ego vehicle trained by attention-DQN showed the lowest collision rate of 14.57\% and the highest success rate of 52.58\%. For the roundabout scenario, similarly, attention-DQN achieved the lowest collision rate of 12.80\% and presented a significant increase in the success rate, from 33.99\% of DQN to 83.92\% of attention-DQN. The freezing rate of the attention-DQN unveiled the lowest value of 3.28\%, which was reduced significantly compared to that of DQN (51.31\%). We also noticed that the performance of multiQ-attention-DQN ranked second, slightly worse than attention-DQN, but very close.

\begin{table}[!ht]
\caption {The testing performance of DQN, single-query and multi-query attention-DQN in intersection and roundabout scenarios.} \label{table_2} 
\begin{adjustbox}{width=\columnwidth,center}
\begin{tabular}{@{}c|cccc@{}}
\toprule
\multirow{2}{*}{\textbf{Scenarios}}    & \multirow{2}{*}{\textbf{\begin{tabular}[c]{@{}c@{}}Metrics\\ mean (std) \%\end{tabular}}} & \multirow{2}{*}{\textbf{DQN}} & \multicolumn{2}{c}{\textbf{Attention-DQN}}   \\ \cmidrule(l){4-5} 
                                       &                                         &                               & \textbf{Default} & \textbf{MultiQ} \\ \midrule
\multirow{3}{*}{\textbf{Intersection}} & \textit{Collision Rate}               & 56.88 (5.39)   & 14.57 (3.76)  & 18.54 (6.40)                 \\
                                       & \textit{Success Rate}                 & 29.23 (4.99)   & 52.58 (6.35)  & 49.18 (6.74)                 \\
                                       & \textit{Freezing Rate}                & 13.89 (3.81)   & 32.85 (6.03)  & 32.28 (8.88)                 \\ \midrule
\multirow{3}{*}{\textbf{Roundabout}}   & \textit{Collision Rate}               & 14.70 (7.24)   & 12.80 (5.30)  & 12.93 (4.73)                 \\
                                       & \textit{Success Rate}                 & 33.99 (5.66)   & 83.92 (4.30)  & 80.90 (8.06)                 \\
                                       & \textit{Freezing Rate}                & 51.31 (10.33)  & 3.28 (3.17)   & 6.17 (9.93)                 
\end{tabular}
\end{adjustbox}
\end{table}

\subsubsection{Our Findings}

The evaluation outcomes of attention-DQN effectively found improved safety concerns of the autonomous vehicle in intersection and roundabout driving scenarios. In the intersection scenario, translation as the attention mechanism helped the ego vehicle pay better attention to the other vehicles in the traffic, thus avoiding collisions and completing the task more efficaciously and successfully. In the roundabout scenario, attention-DQN showed a meagre freezing rate of ego vehicles of only 3.28\% in the test phase, suggesting that the attention mechanism alleviates the problem of robot freezing because the ego vehicle is more confident in moving forward with supplementary attention information to the surrounding vehicles.

\begin{figure}[!htp]
\centering
\includegraphics [width=\columnwidth] {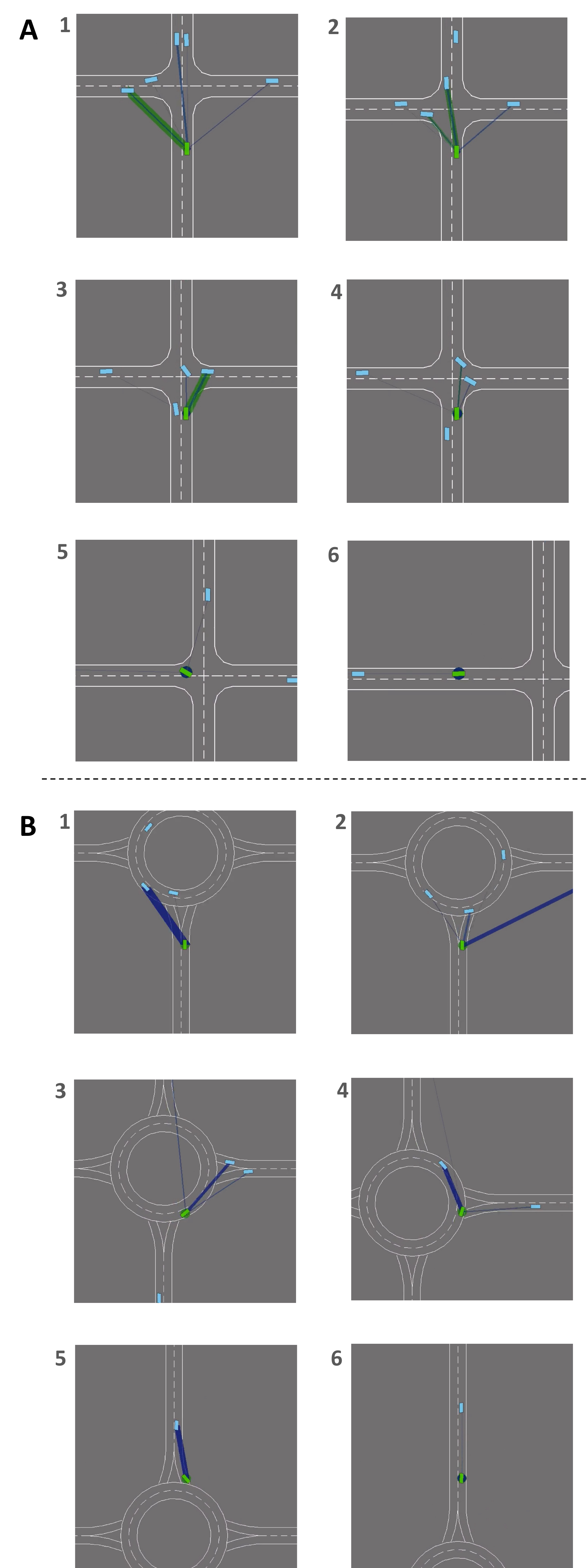}
\caption{A showcase of testing episodes using attention-DQN in (A) intersection and (B) roundabout scenarios. The two attention heads are visualised as green and blue lines connecting the ego vehicle and other surrounding vehicles, and the width of a line corresponds to the weight of the attention. }
\label{fig_exp2_findings}
\end{figure}

In addition, we demonstrated a showcase 
\footnote {A testing demonstration of replay videos is available at \href{https://github.com/czh513/DRL_Self-Awareness-Safety}{GitHub}. } 
in \textbf{Figure \ref{fig_exp2_findings}} about testing episodes using attention-DQN in intersection and roundabout scenarios to further clarify our findings. Here, please note that "attention" is visualised as lines connecting the ego vehicle and other surrounding vehicles. The two different colours (green and blue) represent two attention heads, and the width of the line corresponds to the weight of the attention. The goal of the ego vehicle is to turn left in the intersection scenario (\textbf{Figure \ref{fig_exp2_findings}-A}), and the training attention behaviours of the ego vehicle follow as: (1) the ego vehicle paid attention to the surrounding vehicles from left, front, and right directions; (2) when ego vehicle coming closer to the intersection, the attention to the incoming other vehicles strengthened; (3) the surrounding vehicles from the left to right turn or from the front to left turn at the intersection that were no longer threats to the ego vehicle, so the main attention switched to the surrounding vehicles from the right whose intention was not clear yet;  (4) becasue the vehicles from the right showed intention to turn right and drove almost out of the intersection, the attention on it decreased; (5) the intersection was clear for all surrounding vehicles, so that the attention came back to the ego vehicle itself when the ego vehicle crossed the intersection; (6) the ego vehicle drove on the destination lane with slight attention to the vehicle in front of it to keep distance. Moreover, targetting to the roundabout scenario (\textbf{Figure \ref{fig_exp2_findings}-B}), the ego vehicle needs to learn to cross the roundabout and take the desired exit, and the training attention behaviours of the ego vehicle follow as follows: (1) the ego vehicle paid substantial attention to the surrounding vehicles coming from the left that were likely to collide with the ego vehicle; (2) the ego vehicle waited until the surrounding vehicle from left passed the entry point and switched its primary attention to the surrounding vehicles coming from the following entry; (3) the ego vehicle entered the roundabout and kept its attention on other surrounding vehicles; (4) the attention to the front vehicle strengthened, since the ego vehicle was getting closer to its front vehicle; (5) the ego vehicle exited the roundabout but still kept its attention on the front vehicle to keep the reasonable distance; and (6) the attention switched back to the ego vehicle when it was safe enough.

\section{Discussion}

\begin{figure}[!htp]
\centering
\includegraphics [width=4cm] {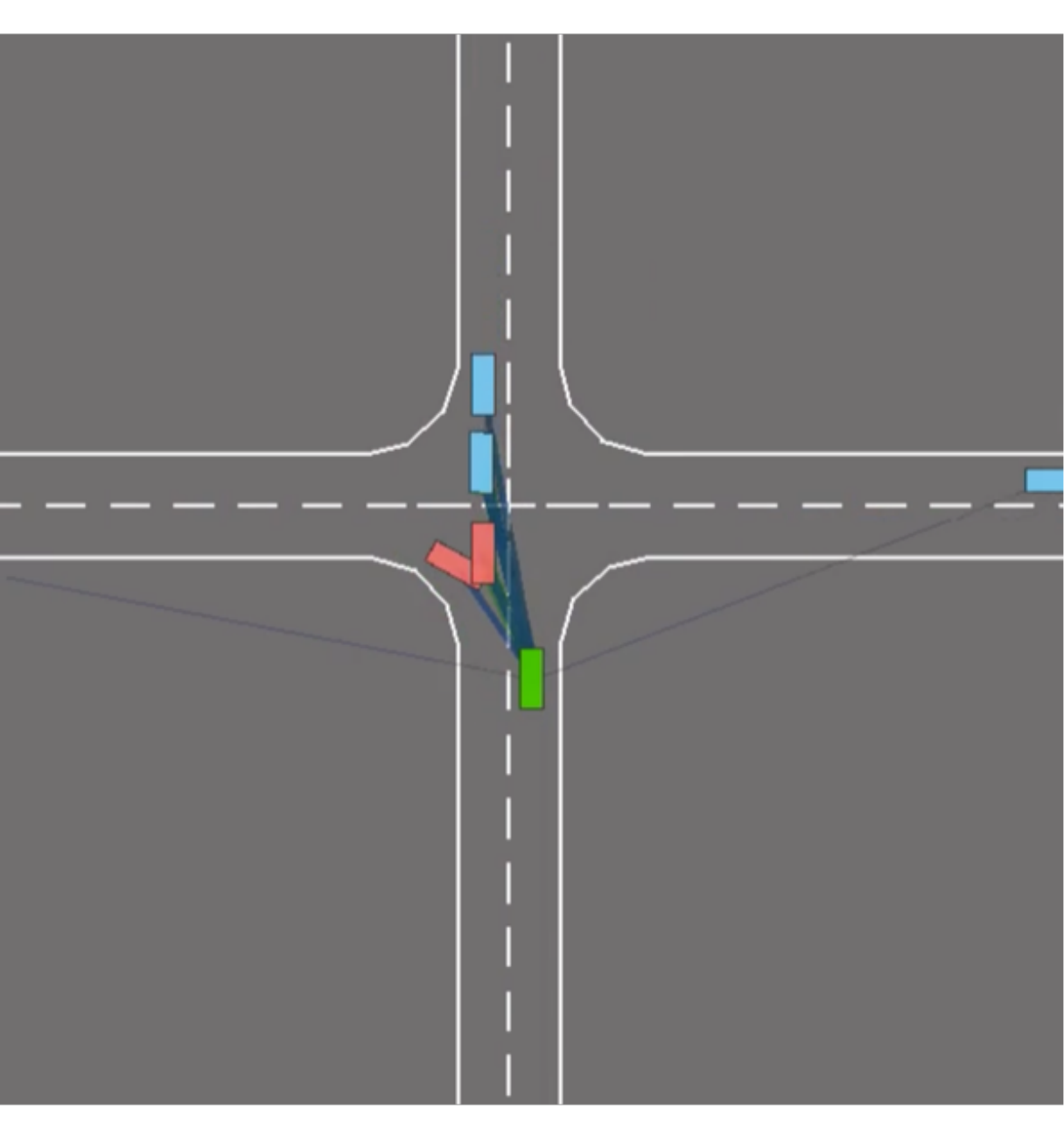}
\caption{A showcase of the frozen vehicle in intersection scenario.}
\label{fig_11}
\end{figure}

Based on our findings, we observed that Attention-DQN lowered the freezing rate of the ego vehicle significantly in the roundabout scenario, from above 50\% freezing rate in both training and testing results to only 3.28\% in testing and 15\% in training. This translates as a significant improvement on the freezing robot problems in the roundabout scenario. One possible reason we speculated that Attention-DQN helped to achieve a low freezing rate is that the attention mechanism enables the ego vehicle to be more confident in moving forward by paying more attention to the surrounding vehicles. Interestingly, the attention mechanism did not have the same effect on the ego vehicle in the intersection scenario. The freezing rates of both training and testing results after the applied attention mechanism were higher than the baseline algorithm. The collision rate did decrease a lot in the intersection scenario, but not all of the decreased percentage effectively transferred to the success rate and some part of the collision rate transferred to the freezing rate. It is unclear in the current stage, but one observation during the experiments was that crashes between other vehicles happened in intersections which sometimes blocked the path of the ego vehicle. In this case, the ego vehicle without attention collided with the crashed vehicles, but with attention, it has had to choose to freeze, as referred in \textbf{Figure \ref{fig_11}}.

For DRL in a complex autonomous driving environment, our research still presented two limitations: The simulated driving environment might exhibit stochastic behaviours that included our testing driving scenarios are complex scenarios populated with vehicles where one vehicle on the road may collide with another. During the experiments, we observed that sometimes there were crashes between other vehicles, which blocked the path of our ego vehicle. As a result, the ego vehicle could not move forward and could only wait till timeout, but it was used to calculate into freezing rate, suggesting that the freezing rate might be influenced by this case. Also, our proposed model is still not entirely safe to be applied to real-world autonomous driving, so it is worth exploring to combine optimisation measurements and more sensing information with DRL trained agents for further safety improvement.

\section{Conclusion}

In this study, we evaluated the safety performance of autonomous driving in a complex road traffic junction environment, such as challenging intersection and roundabout scenarios. We designed four evaluation metrics, including collision rate, success rate, freezing rate, and total reward, from three baseline DRL models (DQN, A2C, and PPO) to the proposed attention-DQN module for self-awareness safety improvement. Our results based on two experiments showed drawbacks of the current DRL models and presented significantly reduced collision rate and freezing rate by introducing the attention mechanism that validated the effectiveness of our proposed attention-DQN module in enhancing the safety concerns of the ego vehicle. Our findings have the potential to contribute to and benefit safe DRL in transportation applications.





\balance
\bibliographystyle{IEEEtran}
\bibliography{IEEEabrv,newref}

\vfill

\end{document}